\pdfoutput=1
\documentclass[11pt,a4paper]{article}

\usepackage[T1]{fontenc}
\usepackage[utf8]{inputenc}
\usepackage{lmodern}
\usepackage[margin=2.5cm]{geometry}
\usepackage{amsmath,amssymb,amsthm,mathtools,bm}
\usepackage{graphicx}
\usepackage{booktabs}
\usepackage{array}
\usepackage[round,authoryear]{natbib}
\usepackage[font=small,labelfont=bf,labelsep=period]{caption}
\usepackage{enumitem}
\usepackage{microtype}
\usepackage{xcolor}
\usepackage{tikz}
\usetikzlibrary{arrows.meta}
\usepackage[hidelinks]{hyperref}

\definecolor{cblue}{HTML}{2A78D6}
\definecolor{corange}{HTML}{EB6834}
\definecolor{cviolet}{HTML}{6250D6}
\definecolor{linegray}{HTML}{52514E}
\definecolor{textgray}{HTML}{52514E}

\setlist{itemsep=2pt,topsep=4pt}
\newtheorem{proposition}{Proposition}

\newcommand{\CKA}{\mathrm{CKA}}
\newcommand{\CKAs}{\mathrm{CKA}^{*}}
\newcommand{\HSIC}{\widehat{\mathrm{HSIC}}}
\newcommand{\Cov}{\operatorname{Cov}}
\newcommand{\E}{\mathbb{E}}
\newcommand{\norm}[1]{\lVert #1 \rVert_{F}}
\newcommand{\ip}[2]{\langle #1, #2 \rangle_{F}}

\title{\Large\textbf{Platonic Representations under Lossy Observations:\\ Reliability-Corrected Estimation of Cross-Modal Alignment}}
\author{Fatih Kutlu\\[2pt]
\small Department of Artificial Intelligence and Robotics, Van Yüzüncü Yıl University, Van, Türkiye\\
\small \texttt{fatihkutlu@yyu.edu.tr}}
\date{}

\begin{document}
\maketitle

\begin{abstract}
\noindent
Claims that models trained on different modalities converge to similar representations are usually tested by measuring the alignment between an image and a single caption of it. A single caption, however, conveys only part of a scene, and different captions of the same scene emphasize different details. This observation noise lowers measured alignment, and differences in how sensitive models are to it masquerade as representational differences. We propose a measurement model based on repeated observations of the same scene. The model factorizes observed linear CKA into the alignment between scene-conditional mean representations ($\CKAs$) and the reliability factors of the two modalities. The reliability factor is defined for a general noise covariance. The classical Spearman--Brown correction is a special case of it and, whenever the noise is not proportional to the scene component, always understates the gain from averaging. All components are estimated with unbiased HSIC U-statistics and a scene-level weighted bootstrap. We test the method in two studies with analysis plans locked in advance, measuring the alignment between DINOv2-g and 16 language models on 3,200 COCO images and 5,000 MSR-VTT video clips. On MSR-VTT, a prediction built from only two captions matched the alignment observed with the mean of 6, 12 and 18 captions on the same clips with a mean absolute error of at most 0.004, whereas the Spearman--Brown prediction fell about 0.01 below the observed values. Reliability differences changed model rankings: a 22-million-parameter sentence encoder led all 7B models under raw measurement but fell behind the Pythia family on MSR-VTT after correction. In the Pythia family, the decline of raw alignment with scale was driven largely by falling text reliability in both datasets, whereas family differences and the scale trend of corrected alignment depended on the dataset. Comparisons of representational convergence cannot be interpreted soundly without accounting for observation reliability.

\end{abstract}

\section{Introduction}

Whether large models trained on different data and with different objectives learn similar internal representations has become an increasingly prominent question in representation learning. \citet{huh2024platonic} answered it with the Platonic representation hypothesis: as models grow and are trained on more diverse tasks, they converge to a shared statistical model of the reality underlying the data. The most striking evidence for the hypothesis is cross-modal alignment. As language models trained only on text become stronger, they produce neighborhood structures on the same image--text pairs that increasingly resemble those of image encoders. The structural similarity found between unaligned vision and language encoders \citep{maniparambil2024vision}, the increase in alignment when exemplars are averaged \citep{he2025seeing}, and a theoretical convergence result for joint-embedding predictive architectures \citep{balestriero2026proving} support this picture.

The measurement basis of the hypothesis has been questioned at the same time. \citet{koepke2026cave} showed that measured alignment drops markedly as the sample grows and in the many-to-many regime, in which each image has several captions. \citet{groger2026aristotelian} found that common similarity metrics are confounded with model size. The common message of these critiques is clear: conclusions about cross-modal convergence depend as much on the measurement setup as on the representations themselves.

This paper focuses on one component of the measurement setup: the fact that observations are lossy. A human caption of an image is not a complete copy of the scene. The annotator mentions some objects and omits others, and adopts a particular viewpoint and choice of words. Five captions of the same image describe the same scene, but each foregrounds a different part of it. The representation that a language model assigns to a single caption therefore carries two components: a component shared by the scene and a deviation specific to that caption. When alignment is measured with a single caption, the second component acts as noise and pulls the measured value down. This is the counterpart, for representational alignment, of the classical attenuation problem, in which the correlation between two noisy measurements is weakened in proportion to their reliability \citep{spearman1904proof}. In neuroscience, the same problem has long been handled with noise ceilings computed from repeated stimulus presentations \citep{nili2014toolbox,schoppe2016measuring,lage2019methods}. In studies of alignment between vision and language models, by contrast, observation noise is generally left uncorrected.

This omission has two consequences. First, the absolute level of alignment is underestimated: even if two models represented a scene in exactly the same way, a measurement based on single captions would not show them as perfectly aligned. Second, and more importantly, model comparisons can be misleading. Language models differ markedly in how consistently they represent different captions of the same scene, and this consistency also changes with scale within a family. A difference in raw alignment then reflects two properties at once: how well the shared component of the scene agrees with the image representation, and how sensitive the model is to caption-specific details. One of our findings shows the size of this confound clearly. A 22-million-parameter sentence encoder showed higher alignment than every 7B language model studied when measured with single captions, yet fell behind the Pythia family on MSR-VTT once observation noise was corrected.

We propose a measurement model based on repeated observations of the same scene. When each scene has several observations in each modality (several captions per image, several frame sets per clip), the covariance of the scene component and the covariance of the observation-specific component can be estimated separately. This separation factorizes linear centered kernel alignment (CKA) into two factors: the alignment between scene-conditional means and the reliability factor of each modality (Figure~\ref{fig:schema}). The decomposition also predicts what the alignment will be when a given number of observations is averaged. This prediction is the testable consequence of the method: a curve built from only two observations can be compared with the alignment measured directly with the mean of more observations.

Our contributions are as follows:
\begin{enumerate}[label=(\roman*)]
  \item A measurement model that factorizes observed linear CKA into the alignment between scene means and the reliability factors of the two modalities. The reliability factor is defined for a general noise covariance and contains the Spearman--Brown correction as a special case. We prove that this correction always understates the gain from averaging whenever the noise is not proportional to the scene component.
  \item Estimators based on unbiased HSIC U-statistics for all components; a bilinearity identity that makes their computation memory-efficient for many observations; a weighted bootstrap that removes the bias caused by duplicated scenes; and a disjoint-observation rule that prevents the inflation caused by shared observations when two models of the same modality are compared.
  \item Validation in two studies with analysis plans locked in advance. On MSR-VTT, the prediction built from two captions matched the alignment observed with the mean of 6, 12 and 18 captions on the same clips with an error far below the prespecified tolerance, whereas the Spearman--Brown prediction deviated in the direction predicted by theory.
  \item A demonstration in two datasets that reliability correction changes model rankings and the interpretation of scale trends. We report separately which findings replicate across the two datasets and which depend on the dataset.
\end{enumerate}

The scope of the study should be stated at the outset. We address not whether representations converge but how convergence should be measured. Corrected alignment measures alignment not with the scene itself but with the human caption expected under a particular captioning protocol. The analyses are restricted to linear CKA and to frozen pretrained encoders.

\section{Related Work}

\paragraph{Representation similarity measures.} Methods based on canonical correlation \citep{raghu2017svcca}, representational similarity analysis \citep{kriegeskorte2008rsa} and centered kernel alignment \citep{cortes2012algorithms,kornblith2019similarity} are widely used to compare neural network representations; comprehensive surveys are given by \citet{klabunde2025similarity} and \citet{sucholutsky2025getting}. Linear CKA scales the Frobenius norm of the cross-covariance of two representations by the norms of their own covariances, and it is the normalized form of the Hilbert--Schmidt independence criterion (HSIC) with linear kernels \citep{gretton2005measuring}. The unbiased U-statistic of HSIC \citep{song2012feature} has been used to compute CKA without bias in minibatches and finite samples \citep{nguyen2021wide}. What these measures capture is itself debated: \citet{ding2021grounding} tested their sensitivity to functionally important changes with statistical tests, and \citet{davari2023reliability} showed that CKA can be overly sensitive to outliers and to relatively small transformations. The problem we address concerns less the metric itself than the reliability of the observations to which it is applied.

\paragraph{Cross-modal convergence.} \citet{huh2024platonic} used mutual $k$-nearest-neighbor overlap as the main measure for cross-modal comparisons and reported that the alignment of language models with image encoders increases with model capability. \citet{maniparambil2024vision} measured a similar structural correspondence with CKA, and \citet{he2025seeing} observed that alignment increases rather than decreases when the embeddings of several exemplars of the same concept are averaged. \citet{balestriero2026proving} proved that joint-embedding predictive architectures trained separately on different modalities learn the same representation up to a rotation when their positive pairs consist of two independent observations of the same scene (good views), and interpreted next-token prediction as joint-embedding prediction with bad views. The same work defined the fraction of an encoder's representation explained by the scene state and showed that the lower bound on the alignment error remaining after the best rotation between two modalities grows as this fraction shrinks. On the critical side, \citet{koepke2026cave} reported that the measure of Huh et al.\ was computed on about a thousand samples and that the increasing trend with scale does not appear for newer models. In the same work, more detailed captions increased alignment, but the gain saturated early and cross-modal agreement remained below the same-modality ceiling. \citet{groger2026aristotelian} showed that common metrics are systematically inflated by model width and depth and proposed a permutation-based calibration. \citet{umar2026snr} showed in a controlled setting that the signal-to-noise ratio and the sample size of the training data govern alignment between networks. These studies focus on factors in the training data, in the metric or in the content of the captions. The problem addressed here is that the observations used at evaluation time are themselves noisy, and that this noise has to be separated statistically from the alignment measure.

\paragraph{Measurement error and noise ceilings.} That the correlation between two measurements is attenuated by measurement error and can be corrected with repeated measurements has been known in psychometrics for over a century \citep{spearman1904proof,lord1968statistical}. The Spearman--Brown formula \citep{spearman1910correlation,brown1910some} predicts from a single reliability coefficient how reliability changes when a measurement is lengthened. In neuroscience, the highest similarity attainable between models and brain responses is assessed with noise ceilings computed from the consistency across repeated presentations \citep{nili2014toolbox,schoppe2016measuring,lage2019methods}. \citet{muzellec2026masked} computed a noise ceiling from split-half repetitions for CKA between recorded neural responses and scaled alignment values by this ceiling. Bias arising from sampling has also been studied: \citet{chun2025estimating} proposed an estimator that corrects CKA for bias due to both input and feature sampling, and \citet{kang2025spectral} showed in a spectral framework that similarity measured with a limited number of neurons is systematically underestimated.

This study carries replicate-based reliability correction over to cross-modal alignment between machine learning models and departs from previous approaches in four respects. First, it defines the reliability factor for a general noise covariance and shows that Spearman--Brown is a special case of this factor that, in general, makes a directional error. Second, it performs the estimation with replicates in both modalities and with unbiased U-statistics for every component. Third, it defines a rule that prevents the inflation caused by shared observations when two models of the same modality are compared. Fourth, it tests the averaging curve predicted by the decomposition directly against observation, with a prespecified tolerance.

\section{Method}

\subsection{Measurement model}
\label{sec:model}

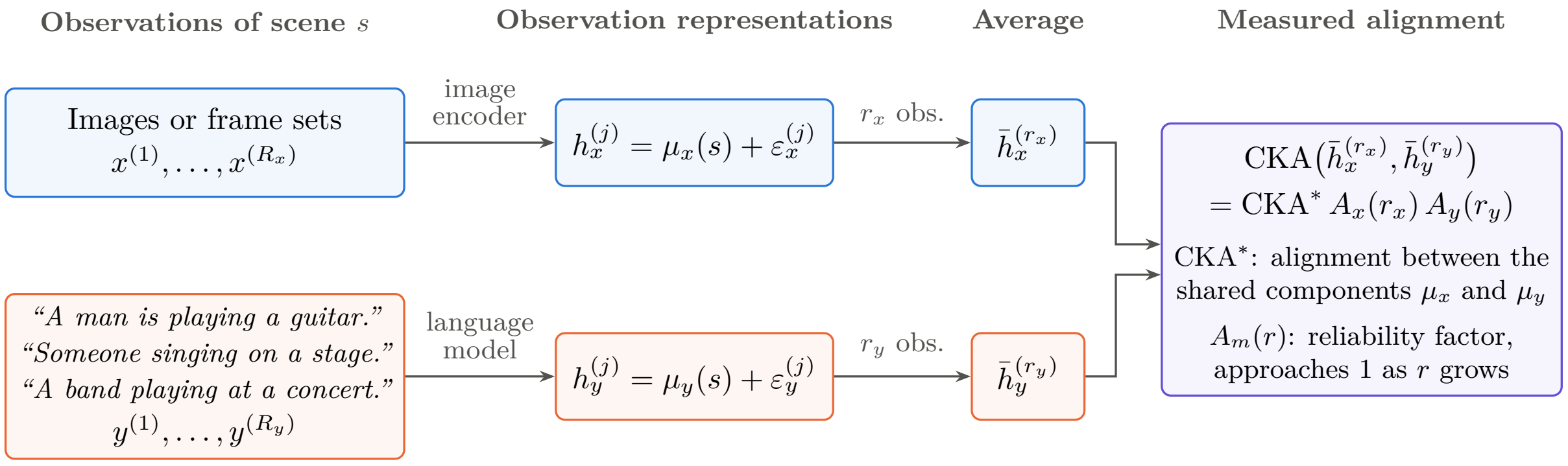
\begin{figure}[t]
\centering
\resizebox{\linewidth}{!}{% Schematic of the measurement model (Figure 1). Included in main.tex via \input{schema}.
\begin{tikzpicture}[
    font=\small,
    box/.style={draw=#1, line width=0.7pt, rounded corners=3pt, fill=#1!6, align=center, inner sep=4pt},
    arr/.style={-{Stealth[length=5pt]}, line width=0.7pt, draw=linegray},
    lab/.style={font=\footnotesize, text=textgray, align=center},
    head/.style={font=\footnotesize\bfseries, text=textgray, align=center},
  ]
  % Column headers
  \node[head] at (2.30,2.75) {Observations of scene $s$};
  \node[head] at (7.95,2.75) {Observation representations};
  \node[head] at (11.80,2.75) {Average};
  \node[head] at (15.65,2.75) {Measured alignment};

  % Observations
  \node[box=cblue, minimum width=4.6cm, minimum height=1.25cm] (gx) at (2.30,1.35)
    {Images or frame sets\\[2pt] $x^{(1)}, \dots, x^{(R_x)}$};
  \node[box=corange, minimum width=4.6cm] (gy) at (2.30,-1.35)
    {{\footnotesize\itshape ``A man is playing a guitar.''}\\
     {\footnotesize\itshape ``Someone singing on a stage.''}\\
     {\footnotesize\itshape ``A band playing at a concert.''}\\[2pt]
     $y^{(1)}, \dots, y^{(R_y)}$};

  % Representations
  \node[box=cblue, minimum width=3.2cm, minimum height=1.0cm] (hx) at (7.95,1.35)
    {$h_x^{(j)} = \mu_x(s) + \varepsilon_x^{(j)}$};
  \node[box=corange, minimum width=3.2cm, minimum height=1.0cm] (hy) at (7.95,-1.35)
    {$h_y^{(j)} = \mu_y(s) + \varepsilon_y^{(j)}$};

  % Means
  \node[box=cblue, minimum width=1.3cm, minimum height=1.0cm] (mx) at (11.80,1.35) {$\bar h_x^{(r_x)}$};
  \node[box=corange, minimum width=1.3cm, minimum height=1.0cm] (my) at (11.80,-1.35) {$\bar h_y^{(r_y)}$};

  % Decomposition
  \node[box=cviolet, minimum width=4.4cm] (d) at (15.65,0)
    {$\mathrm{CKA}\big(\bar h_x^{(r_x)}, \bar h_y^{(r_y)}\big)$\\[3pt]
     $= \mathrm{CKA}^{*}\, A_x(r_x)\, A_y(r_y)$\\[5pt]
     {\footnotesize $\mathrm{CKA}^{*}$: alignment between the}\\[-1pt]
     {\footnotesize shared components $\mu_x$ and $\mu_y$}\\[3pt]
     {\footnotesize $A_m(r)$: reliability factor,}\\[-1pt]
     {\footnotesize approaches 1 as $r$ grows}};

  % Arrows
  \draw[arr] (gx.east) -- node[lab, above=2pt] {image\\[-2pt]encoder} (hx.west);
  \draw[arr] (gy.east |- hy.west) -- node[lab, above=2pt] {language\\[-2pt]model} (hy.west);
  \draw[arr] (hx.east) -- node[lab, above=2pt] {$r_x$ obs.} (mx.west);
  \draw[arr] (hy.east) -- node[lab, above=2pt] {$r_y$ obs.} (my.west);
  \draw[arr] (mx.east) -- ++(0.35,0) |- ([yshift=5pt]d.west);
  \draw[arr] (my.east) -- ++(0.35,0) |- ([yshift=-5pt]d.west);
\end{tikzpicture}}
\caption{The measurement model. Each scene has several observations in each modality. The representation an encoder assigns to an observation is the sum of a component shared by the scene, $\mu_m(s)$, and an observation-specific deviation. The CKA between observation means factorizes into the alignment between the shared components ($\CKAs$) and the reliability factors of the two modalities. As the number of averaged observations grows, the reliability factors approach 1 and the observed CKA approaches $\CKAs$.}
\label{fig:schema}
\end{figure}

For a scene $s$ and a modality $m \in \{x, y\}$, write the representation that an encoder produces from the $j$-th observation of the scene as
\begin{equation}
  h_m^{(j)} = \mu_m(s) + \varepsilon_m^{(j)}, \qquad \E\!\left[\varepsilon_m^{(j)} \mid s\right] = 0 .
  \label{eq:model}
\end{equation}
Here $\mu_m(s) = \E[h_m^{(j)} \mid s]$ is the scene-conditional mean representation and $\varepsilon_m^{(j)}$ is the observation-specific deviation (Figure~\ref{fig:schema}). $x$ denotes the image or video side and $y$ the text side. On the text side, observations are different human captions of the same scene; on the video side, they are different frame sets taken from the same clip, which we call views. On COCO there is a single observation on the image side: since every encoder sees the same image file, $\varepsilon_x = 0$.

We denote the scene-level covariance by $B_m = \Cov(\mu_m(s))$, the average observation noise by $W_m = \E_s[\Cov(h_m^{(j)} \mid s)]$, and the total covariance of a single observation by $V_m = B_m + W_m$. The cross-modal scene covariance is $C = \Cov(\mu_x(s), \mu_y(s))$. The model rests on two assumptions.
\begin{description}[font=\normalfont\itshape, leftmargin=1.2em]
  \item[A1 (within-modality exchangeability).] Given the scene, observations of the same modality come from the same distribution and their observation-specific deviations are uncorrelated: $\E[\varepsilon_m^{(j)} \varepsilon_m^{(k)\top} \mid s] = 0$ for $j \neq k$.
  \item[A2 (cross-modal uncorrelatedness).] Given the scene, the observation-specific deviations of the two modalities are uncorrelated: $\E[\varepsilon_x^{(j)} \varepsilon_y^{(k)\top} \mid s] = 0$.
\end{description}
Under A1, the mean $\bar h_m^{(r)}$ of $r$ observations satisfies $\Cov(\bar h_m^{(r)}) = B_m + W_m / r$. Under A2, $\Cov(\bar h_x^{(r_x)}, \bar h_y^{(r_y)}) = C$; the cross-covariance does not depend on the number of observations.

The justification for A1 is the data collection protocol. If the captions were written independently by different people and their order of use is randomized, the captions are exchangeable. Shared tendencies of annotators, for instance a protocol that encourages mentioning certain objects, are then absorbed into the scene component rather than into the noise. Hence $\mu_y(s)$ represents not the scene itself but the representation of the caption expected under a particular captioning protocol. A2 holds trivially on COCO because $\varepsilon_x = 0$. On MSR-VTT, views are selected from a fixed temporal grid independently of the captions, so A2 is a reasonable assumption; it is, however, not tested directly.

\subsection{Decomposition of observed alignment}

For two random vectors $u$ and $v$, population linear CKA is defined as
\begin{equation}
  \CKA(u, v) = \frac{\norm{\Cov(u, v)}^{2}}{\norm{\Cov(u)} \, \norm{\Cov(v)}}
\end{equation}
\citep{kornblith2019similarity}. Under this definition and A1--A2, the following decomposition is immediate.

\begin{proposition}
\label{prop:decomposition}
Under A1 and A2, the linear CKA between the means of $r_x$ image observations and $r_y$ text observations factorizes as
\begin{equation}
  \CKA\big(\bar h_x^{(r_x)}, \bar h_y^{(r_y)}\big) = \CKAs \, A_x(r_x) \, A_y(r_y),
  \label{eq:decomposition}
\end{equation}
where
\begin{equation}
  \CKAs = \frac{\norm{C}^{2}}{\norm{B_x} \, \norm{B_y}} = \CKA\big(\mu_x(s), \mu_y(s)\big), \qquad
  A_m(r) = \frac{\norm{B_m}}{\norm{B_m + W_m / r}} .
\end{equation}
\end{proposition}
\begin{proof}
Under A2 the cross-covariance in the numerator equals $C$, and under A1 the covariances in the denominator equal $B_m + W_m / r_m$. Multiplying and dividing the fraction by $\norm{B_x}\norm{B_y}$ gives the expression.
\end{proof}

$\CKAs$ is the alignment between the scene-conditional means of the two encoders; we call it the corrected alignment. $A_m(r)$ is the reliability factor of a measurement that uses the mean of $r$ observations. As long as $B_m \neq 0$, positive semidefiniteness of $B_m$ and $W_m$ gives $\ip{B_m}{W_m} \geq 0$ and hence $0 < A_m(r) \leq 1$. $A_m(r)$ increases with the number of observations and tends to 1 as $r \to \infty$. The decomposition shows that low raw alignment can have two distinct causes: the scene means may genuinely be weakly aligned, or in at least one modality a single observation may not carry the scene reliably. The decomposition is additive on the log scale, $\log \CKA = \log \CKAs + \log A_x + \log A_y$, which makes it possible to attribute a scale trend to its components.

We report three quantities. \emph{Raw alignment} is the CKA between a single image or view and a single caption, and corresponds to $\CKAs A_x(1) A_y(1)$. \emph{Averaged alignment} is the CKA computed directly between the means of all available observations: on COCO between the image and the mean of five captions, on MSR-VTT between the mean of four views and the mean of 20 captions. \emph{Corrected alignment} is $\CKAs$. The first two are observed directly. The third rests on the measurement model and corresponds to the value reached as the number of averaged observations grows without bound.

\subsection{The form of the reliability factor and Spearman--Brown}

The inverse square of the reliability factor is a quadratic polynomial in $1/r$:
\begin{equation}
  \frac{1}{A_m(r)^{2}} = 1 + \frac{2\kappa_1}{r} + \frac{\kappa_2}{r^{2}}, \qquad
  \kappa_1 = \frac{\ip{B_m}{W_m}}{\norm{B_m}^{2}}, \quad \kappa_2 = \frac{\norm{W_m}^{2}}{\norm{B_m}^{2}} .
  \label{eq:polynomial}
\end{equation}
If the noise covariance is proportional to the scene covariance ($W_m = \lambda B_m$, $\lambda \geq 0$), then $\kappa_1 = \lambda$ and $\kappa_2 = \lambda^2$, and the factor reduces to the classical Spearman--Brown form:
\begin{equation}
  A_m(r) = \frac{r \rho_m}{1 + (r - 1)\rho_m}, \qquad \rho_m = A_m(1) .
\end{equation}
In general, the noise need not concentrate in the same directions as the scene component. Differences between captions, for example, may accumulate in directions related to word choice and sentence structure rather than to the content of the scene. In which direction Spearman--Brown then mispredicts the gain from averaging is settled by the following proposition.

\begin{proposition}
\label{prop:sb}
Let $B_m \neq 0$ and $\rho_m = A_m(1)$, and define $A^{\mathrm{SB}}_m(r) = r\rho_m / (1 + (r-1)\rho_m)$. Then $A_m(r) \geq A^{\mathrm{SB}}_m(r)$ for every $r \geq 1$. For $r > 1$, equality holds if and only if $W_m$ is proportional to $B_m$.
\end{proposition}
\begin{proof}
Let $T = 2\kappa_1 + \kappa_2$; since $1/\rho_m^2 = 1 + T$, $T$ is determined by $\rho_m$. Under the assumption $\kappa_2 = \kappa_1^2$, Spearman--Brown uses the value $\lambda_{\mathrm{SB}} = \sqrt{1 + T} - 1 \geq 0$ that solves $\lambda^2 + 2\lambda = T$; indeed, with this value \eqref{eq:polynomial} gives $1/A^{\mathrm{SB}}_m(r)^2 = (1 + \lambda_{\mathrm{SB}}/r)^2$. The Cauchy--Schwarz inequality gives $\ip{B_m}{W_m} \leq \norm{B_m}\norm{W_m}$, that is, $\kappa_1^2 \leq \kappa_2 = T - 2\kappa_1$. Hence $\kappa_1^2 + 2\kappa_1 - T \leq 0$, and since $\kappa_1 \geq 0$, $\kappa_1 \leq \lambda_{\mathrm{SB}}$. Using \eqref{eq:polynomial},
\[
  \frac{1}{A_m(r)^{2}} - \frac{1}{A^{\mathrm{SB}}_m(r)^{2}} = 2(\kappa_1 - \lambda_{\mathrm{SB}})\left(\frac{1}{r} - \frac{1}{r^{2}}\right) \leq 0 ,
\]
since $1/r - 1/r^2 \geq 0$ for $r \geq 1$. For $r > 1$, equality holds when $\kappa_1 = \lambda_{\mathrm{SB}}$, that is, in the equality case of the Cauchy--Schwarz inequality, which corresponds to $W_m = \lambda B_m$.
\end{proof}

The intuition behind the proposition is as follows. The same single-observation reliability can arise both when the noise concentrates in the same directions as the scene component and when it spreads into directions in which the scene variance is small. In the second case $\kappa_1$ is small and $\kappa_2$ is large. A larger share of the reliability deficit then falls on the term that decays as $1/r^2$, and averaging increases reliability faster than Spearman--Brown assumes. This directional prediction is compared with direct observation in Section~\ref{sec:validation}.

\subsection{Unbiased estimation from repeated observations}

Collect the representations of the $j$-th observations of $n$ scenes in one modality in the rows of a matrix $H^{(j)}_m \in \mathbb{R}^{n \times d_m}$, and let $K^{(j)}_m = H^{(j)}_m H^{(j)\top}_m$ be its linear Gram matrix. The Gram matrix obtained from the mean of $r$ observations is denoted $\bar K^{(r)}_m$, and that of the mean of all observations $\bar K_m$. For Gram matrices with zeroed diagonals $\tilde K$ and $\tilde L$, the unbiased U-statistic of HSIC with linear kernels is
\begin{equation}
  \HSIC(K, L) = \frac{1}{n(n-3)} \left[ \operatorname{tr}(\tilde K \tilde L) + \frac{\bm{1}^{\top}\tilde K \bm{1} \; \bm{1}^{\top}\tilde L \bm{1}}{(n-1)(n-2)} - \frac{2}{n-2} \, \bm{1}^{\top}\tilde K \tilde L \bm{1} \right] .
  \label{eq:hsic}
\end{equation}
When scenes are sampled independently, this statistic is unbiased for the squared Frobenius norm of the cross-covariance between the representations that generate $K$ and $L$ \citep{song2012feature}. With $R_m \geq 2$ observations per scene in each modality, the components of the measurement model are estimated as follows:
\begin{itemize}
  \item $\norm{B_m}^{2}$: under A1, the cross-covariance between two different observations of the same scene equals $B_m$; hence $\hat b_m = \frac{1}{R_m(R_m-1)} \sum_{j \neq k} \HSIC(K^{(j)}_m, K^{(k)}_m)$.
  \item $\norm{V_m}^{2}$: $\hat v_m = \frac{1}{R_m} \sum_j \HSIC(K^{(j)}_m, K^{(j)}_m)$.
  \item $\norm{C}^{2}$: under A2, $\hat c = \HSIC(\bar K_x, \bar K_y)$.
\end{itemize}
From these we obtain $\widehat{\CKAs} = \hat c / \sqrt{\hat b_x \hat b_y}$, $\hat A_m(1) = \sqrt{\hat b_m / \hat v_m}$ and $\hat A_m(r) = \sqrt{\hat b_m / \HSIC(\bar K^{(r)}_m, \bar K^{(r)}_m)}$. When $\hat b_m \leq 0$, these ratios are treated as undefined. On COCO, $\HSIC(K_x, K_x)$ is used in place of $\hat b_x$ on the image side, and $A_x = 1$.

As the number of observations grows, computing $\hat b_m$ directly requires $R_m(R_m - 1)$ cross terms. Since \eqref{eq:hsic} is linear in each of its two Gram-matrix arguments, with $K_{\Sigma} = \sum_j K^{(j)}_m$ one can write
\begin{equation}
  \hat b_m = \frac{\HSIC(K_{\Sigma}, K_{\Sigma}) - \sum_j \HSIC(K^{(j)}_m, K^{(j)}_m)}{R_m(R_m - 1)} .
\end{equation}
Thus $R_m + 1$ HSIC evaluations suffice. The Gram matrices are formed one at a time and accumulated, so that only their sum is kept in memory.

\subsection{Predicting the averaging curve}
\label{sec:prediction}

The decomposition predicts what the observed alignment will be when a given number of captions is averaged. Keep the image or video side fixed at the mean of all its observations ($\bar K_x$), and use only the first two captions on the text side:
\[
  \hat b = \HSIC(K^{(1)}_y, K^{(2)}_y), \quad
  \hat v = \tfrac{1}{2}\big[\HSIC(K^{(1)}_y, K^{(1)}_y) + \HSIC(K^{(2)}_y, K^{(2)}_y)\big], \quad
  \hat w = \tfrac{1}{4}\HSIC(D, D),
\]
\[
  \hat c_{12} = \HSIC(\bar K_x, \bar K^{(2)}_y), \qquad \hat h_x = \HSIC(\bar K_x, \bar K_x).
\]
Here $D$ is the Gram matrix obtained from the difference of the two caption representations ($H^{(1)}_y - H^{(2)}_y$). Since the covariance of the difference of two observations is $2W_y$, $\hat w$ estimates $\norm{W_y}^2$. From the identities $\norm{B + W/r}^2 = \norm{B}^2 + 2\ip{B}{W}/r + \norm{W}^2/r^2$ and $\norm{V}^2 = \norm{B}^2 + 2\ip{B}{W} + \norm{W}^2$, the covariance norm of the mean of $r$ captions is estimated by
\begin{equation}
  \hat S(r) = \hat b + \frac{\hat v - \hat b - \hat w}{r} + \frac{\hat w}{r^{2}} .
\end{equation}
With this side held fixed, the matrix prediction of the alignment measured with the mean of $r$ captions is
\begin{equation}
  p_{\mathrm{M}}(r) = \frac{\hat c_{12}}{\sqrt{\hat h_x \, \hat S(r)}} .
  \label{eq:prediction}
\end{equation}
The Spearman--Brown prediction is computed from the same components with $\hat\rho = \sqrt{\hat b / \hat v}$ as $p_{\mathrm{SB}}(r) = \hat c_{12} / \sqrt{\hat h_x \hat b} \cdot r\hat\rho / (1 + (r-1)\hat\rho)$. With $o(r)$ the value observed directly with the mean of $r$ captions, the prediction error is $e(r) = o(r) - p(r)$; a positive error means that the prediction falls below the observation.

It should be stated explicitly what the prediction tests. As long as the captions are exchangeable, the cross-covariance between any two captions is the same. The covariance of the mean of $r$ captions is then a linear function of $1/r$, and the form above holds exactly at the population level. What is tested is therefore not the functional form. It is whether components estimated from only two observations are precise enough in finite samples to be carried over to larger values of $r$, and whether the scalar Spearman--Brown form suffices for this extrapolation. A prediction that holds over the observable range also supports interpreting $\CKAs$ as the limit of averaging.

\subsection{Comparing models of the same modality}
\label{sec:disjoint}

When two models of the same modality are compared (two language models or two video encoders), it is common practice to give both models the same observation. When two models read the same caption, however, the content specific to that caption appears in both representations, and the observation-specific deviations become correlated across the models. Within-modality alignment is therefore artificially inflated and cannot be compared on an equal footing with cross-modal pairs that share no observation. In this study, two models of the same modality are compared on disjoint sets of observations of each scene (the disjoint-observation rule). The observations are split into two disjoint halves. Writing $\bar K_{\!A}^{[1]}$ for the Gram matrix that model $A$ produces from the mean of the first half, the corrected alignment between the two models is computed as
\begin{equation}
  \widehat{\CKAs}_{AB} = \frac{\tfrac{1}{2}\big[\HSIC(\bar K_{\!A}^{[1]}, \bar K_{\!B}^{[2]}) + \HSIC(\bar K_{\!A}^{[2]}, \bar K_{\!B}^{[1]})\big]}{\sqrt{\hat b_A \, \hat b_B}} .
\end{equation}
The halves are observations 1--2 and 3--4 for COCO captions, 1--10 and 11--20 for MSR-VTT captions, and 1--2 and 3--4 for MSR-VTT views. On COCO, since image encoders see the same image and $\varepsilon_x = 0$, image--image comparisons use raw CKA. The effect of the rule is measured in Section~\ref{sec:gap}.

\subsection{Uncertainty}

Uncertainty is computed by resampling scenes. A naive bootstrap that duplicates rows counts two copies of the same scene as different scenes. This reintroduces the same-scene terms that \eqref{eq:hsic} excludes by zeroing the diagonal, and it biases HSIC upward. Each bootstrap draw is therefore represented by a multiplicity vector $\bm{\omega} \in \mathbb{N}^{n}$ ($\sum_i \omega_i = n$), and the U-statistic is computed in weighted form:
\begin{equation}
  \HSIC_{\bm{\omega}}(K, L) = \frac{1}{n(n-3)}\left[ t_1 + \frac{t_2}{(n-1)(n-2)} - \frac{2\,t_3}{n-2} \right],
\end{equation}
where $t_1 = \bm{\omega}^{\top}(\tilde K \circ \tilde L) \bm{\omega}$, $t_2 = (\bm{\omega}^{\top}\tilde K \bm{\omega})(\bm{\omega}^{\top}\tilde L \bm{\omega})$ and $t_3 = \sum_i \omega_i (\tilde K \bm{\omega})_i (\tilde L \bm{\omega})_i$. For $\bm{\omega} = \bm{1}$ the expression reduces to \eqref{eq:hsic}, and kernel entries between copies of the same scene are treated as zero. In unit tests released with the code, the bias of the naive bootstrap reached about 180\%, whereas that of the weighted form stayed below 2\%. The same 2,000 bootstrap draws are used for all model pairs, so that comparisons are paired at the level of draws \citep{efron1993introduction}. Intervals are 95\% percentile intervals. Draws with $\hat b \leq 0$ are treated as undefined, and statistics with more than 10\% undefined draws are not interpreted. No undefined draws occurred for any reported statistic.

\section{Experimental Setup}

\subsection{Data}

\paragraph{COCO.} All 5,000 images of the COCO val2017 split have at least five captions \citep{lin2014coco,chen2015coco}. The captions of each image were sorted by their identifiers, shuffled with an image-specific seed, and the first five were kept. Of the images, 1,000 were reserved for the pilot; of the remaining 4,000, 800 were used only for selecting language-model layers and 3,200 for the tests. In the K1 test, which evaluates the averaging prediction, the test images were split into two disjoint halves of 1,600 images.

\paragraph{MSR-VTT.} MSR-VTT \citep{xu2016msrvtt} contains about 20 human captions per clip. We used the training-9k split from a pinned revision of the \texttt{friedrichor/MSR-VTT} repository on Hugging Face; all 9,000 clips in this split have at least 20 captions. Since no model is trained, using the training split raises no leakage concern. The clips were ordered once with a seed; the first 300 decodable clips were used for the pilot, and the next 5,000 decodable clips not examined by the pilot for the full stage. No clip failed to decode in the full stage. The captions of each clip were shuffled with a seed and the first 20 were kept. Exact duplicate captions were not removed: different annotators writing the same sentence is consistent with independent draws from the caption distribution of the clip, and removing such captions would alter that distribution.

On the video side, each clip was divided into eight equal temporal segments, and the frame at the midpoint of each segment was taken. View $j$ ($j = 1, \dots, 4$) is the mean of the features of frames $j$ and $j+4$. Each view thus covers both the first and the second half of the clip, and the four views are interchangeable. On MSR-VTT, raw alignment is the mean of the CKA values of the four (view $j$, caption $j$) pairs.

\subsection{Models and features}

Ten frozen encoders were used on the image side: the small, base, large and giant versions of DINOv2 \citep{oquab2024dinov2}, the base and large versions of the masked autoencoder (MAE) \citep{he2022masked}, ViT-B/16 and ViT-L/16 trained with supervision on ImageNet-21k \citep{dosovitskiy2021image}, and CLIP ViT-B/16 and ViT-L/14 \citep{radford2021learning}. Because they are trained with language supervision, the CLIP encoders were excluded from the tests and from the within-modality reference panel; their results appear only in the released outputs. The main image encoder is DINOv2-g. Features are the mean of the patch tokens of the last layer; the class token (CLS) was also stored for a sensitivity analysis. Masking was disabled for the MAE models.

Sixteen models were used on the language side: five models from the Pythia family (160M--6.9B) \citep{biderman2023pythia}, four from the Qwen2.5 family (0.5B--7B) \citep{qwen2024qwen25}, five from the BLOOM family (560M--7.1B) \citep{bigscience2024bloom}, DistilBERT \citep{sanh2019distilbert}, and the all-MiniLM-L6-v2 sentence encoder \citep{wang2020minilm,reimers2019sentence}. The largest members of the three decoder families (about 7 billion parameters; hereafter the 7B models) are used in the main analyses. The representation is the mean of the non-padding tokens at the selected layer; captions are truncated to at most 64 tokens. The layer was chosen by a prespecified rule among candidates at fractions $\{0.25, 0.375, \dots, 1\}$ of the total depth, as the layer giving the highest raw CKA with DINOv2-g on the COCO selection images. The rule selected the last layer for every decoder model and the fifth of six layers for DistilBERT. Layers were not reselected on MSR-VTT; the layers selected on COCO and the same model revisions were used. Image encoders were run in 32-bit precision and language models in bfloat16. All computations were run on a single consumer GPU with less than 16 GB of memory; part of the layers of the 7B models was executed in CPU memory.

\subsection{Locked analysis plans and prespecified tests}

Both studies followed the same workflow: pilot, power and simulation analysis, freezing of the analysis plan, and full stage. The analysis plan is a YAML file containing the hypotheses, test statistics, decision rules, tolerances, layer rule, model list and seed. When the plan was frozen, its SHA-256 hash was recorded. The full-stage code refuses to run if the plan has not been frozen, if the hash does not match, or if the hash of the pilot manifest differs from the one written in the plan. Pilot and full-stage scenes do not overlap. The MSR-VTT plan also contains the hashes of the COCO layer-selection and model-revision files. The plans were not uploaded to a public registry but locked locally; we therefore speak of analysis plans locked in advance rather than of preregistration \citep[cf.][]{nosek2018preregistration}. The lock files, their hashes and the pilot outputs are released with the code. A large language model (Claude, Anthropic) assisted in planning the analyses and writing the analysis code; the author reviewed all code, ran all experiments and checked the outputs with the unit tests and hash checks released with the code.

Changes made before freezing in response to pilot results were recorded in the plan files. On COCO, the within-modality reference was changed from the panel mean to the highest value from a different family, because in the pilot the MAE models aligned with the other image encoders only at about 0.05--0.09, which made the mean meaningless. In addition, the K1 test was restricted to the pairs of the main image encoder, and the tolerance was set to $\tau = 0.03$. The first run of the MSR-VTT pilot removed exact duplicate captions. Because this rule reduced the number of eligible clips to 3,043 and is also statistically wrong for the reason given above, it was removed before freezing and the pilot was rerun.

Tests are referred to by their names in the lock files. K1 and M1 test the averaging prediction, H1 and H2 the two hypotheses on COCO, and M2 the directional replication of the COCO findings on MSR-VTT. All tests use the main image encoder DINOv2-g.

\paragraph{COCO tests.}
\begin{itemize}
  \item \textbf{K1 (prediction validity).} The matrix prediction \eqref{eq:prediction} built from two captions on the calibration half is compared with the CKA observed on the evaluation half with the mean of $r \in \{3, 4, 5\}$ randomly selected captions. The statistic is the mean $|e(r)|$ over the 16 language models. The test is passed if the upper end of the interval lies below $\tau = 0.03$ for all three values of $r$. If the test is not passed, $\CKAs$ is interpreted not as the limit of averaging but as a model-based estimate of the alignment between the shared components of the captions.
  \item \textbf{H1 (residual gap).} $G$ is the difference between the highest within-modality alignment of DINOv2-g with an encoder from a different family and the mean corrected alignment of DINOv2-g with the three 7B language models. With $\delta = 0.05$, the outcome is a genuine gap if the lower end of the interval lies above $\delta$; equivalence if the interval lies entirely within $(-\delta, \delta)$; cross-modal alignment exceeding within-modality alignment if the upper end lies below $-\delta$; and inconclusive otherwise.
  \item \textbf{H2 (decomposition of the scale trend).} In each decoder family, least-squares slopes of raw CKA, $\log A_y(1)$ and $\log \CKAs$ against the base-10 logarithm of the parameter count are computed. Interpretation of the decomposition was made conditional on the raw slope being positive (lower end of the interval above zero) in at least two families.
\end{itemize}

\paragraph{MSR-VTT tests.} The MSR-VTT study aims to close two points left open by COCO: the precision of the averaging prediction, and whether the COCO findings replicate on a second dataset.
\begin{itemize}
  \item \textbf{M1 (testing the prediction against direct observation).} For each language model, the prediction is built from the first two captions only. For the observation, $r \in \{6, 12, 18\}$ captions are drawn at random with a seed from the remaining 18 captions of each clip, and $o(r)$ is computed from their mean. Unlike in COCO, prediction and observation use the same clips and the same bootstrap draws, which removes the sampling difference between two halves. The statistic and the decision rule are the same as for K1. The number of clips was to be chosen as the smallest candidate in $\{3{,}000, 4{,}000, 5{,}000\}$ for which the upper bound projected from the pilot errors fell below $\tau/2$. Since no candidate met this condition, $n = 5{,}000$ was used, as the plan required.
  \item \textbf{M2 (directional replication of the COCO findings).} Three directional predictions were defined. M2a: the raw CKA of MiniLM-L6 exceeds that of each of the three 7B models. M2b: the difference between the Pythia and Qwen2.5 family means shrinks when moving from raw to averaged alignment. M2c: within the Pythia family, raw CKA decreases with scale while $\log \CKAs$ increases. The outcome of each prediction is recorded as replicated, reversed (interval entirely in the opposite direction) or inconclusive. For M2c, the outcome is recorded as partial if only one of the two conditions holds.
\end{itemize}

\section{Results}

All prespecified tests are summarized in Table~\ref{tab:tests}. We present the findings below by the question they answer rather than by dataset. Analyses that were not prespecified are labeled exploratory.

\begin{table}[t]
\centering
\caption{Results of the prespecified tests. $e(r)$ is the prediction error. For K1 and M1, the decision rule requires the upper end of the interval to lie below $\tau = 0.03$ for all three values of $r$. Slopes are per tenfold increase in parameter count; $\log$ is the natural logarithm. Square brackets give 95\% bootstrap intervals.}
\label{tab:tests}
\small
\renewcommand{\arraystretch}{1.12}
\begin{tabular}{@{}l >{\raggedright\arraybackslash}p{6.0cm} l >{\raggedright\arraybackslash}p{2.6cm}@{}}
\toprule
Test & Statistic (expected direction) & Estimate [95\% interval] & Decision \\
\midrule
\multicolumn{4}{@{}l}{\textit{COCO (3,200 test images)}} \\
K1 & Mean $|e(r)|$, $r = 3$ & 0.009 [0.006, 0.030] & \\
   & Mean $|e(r)|$, $r = 4$ & 0.014 [0.006, 0.034] & \\
   & Mean $|e(r)|$, $r = 5$ & 0.014 [0.006, 0.035] & Not passed \\
H1 & Residual gap $G$ ($\delta = 0.05$) & 0.058 [0.045, 0.072] & Inconclusive \\
H2 & Raw slope, BLOOM ($> 0$) & 0.041 [0.035, 0.047] & \\
   & Raw slope, Qwen2.5 ($> 0$) & 0.004 [0.0001, 0.007] & \\
   & Raw slope, Pythia ($> 0$) & $-0.022$ [$-0.024$, $-0.019$] & Precondition met \\
\addlinespace
\multicolumn{4}{@{}l}{\textit{MSR-VTT (5,000 clips)}} \\
M1 & Mean $|e(r)|$, $r = 6$ & 0.003 [0.003, 0.010] & \\
   & Mean $|e(r)|$, $r = 12$ & 0.004 [0.004, 0.013] & \\
   & Mean $|e(r)|$, $r = 18$ & 0.004 [0.004, 0.014] & Passed \\
M2a & Raw difference, MiniLM-L6 minus best 7B model ($> 0$) & 0.097 [0.092, 0.103] & Replicated \\
M2b & Shrinkage of the Pythia--Qwen2.5 difference: raw minus averaged difference ($> 0$) & $-0.010$ [$-0.015$, $-0.005$] & Reversed \\
M2c & Raw slope, Pythia ($< 0$) & $-0.016$ [$-0.018$, $-0.014$] & \\
    & $\log \CKAs$ slope, Pythia ($> 0$) & $-0.005$ [$-0.009$, $-0.0001$] & Partial \\
\bottomrule
\end{tabular}
\end{table}

\subsection{Validating the averaging prediction}
\label{sec:validation}

\paragraph{Simulation.} The estimators were tested during the COCO pilot on 48-dimensional Gaussian data generated with known parameters. The 24 settings in which the noise is uncorrelated across modalities cover three sample sizes ($n \in \{500, 1{,}000, 2{,}000\}$) and different single-caption reliabilities ($A_y(1) \in \{0.3, 0.5, 0.8\}$). In 18 of these settings only the text side is noisy, with a proportional or non-proportional noise covariance; in six, both sides are noisy. Each setting was run with 200 replications and 500 bootstrap draws. For $\CKAs$, whose true value is 0.102, the largest absolute bias was 0.0012, and the coverage of the 95\% intervals was 0.935--0.990. Under proportional noise, the matrix and Spearman--Brown predictions gave nearly identical errors, differing by less than 0.0001. Under non-proportional noise, the mean absolute error of Spearman--Brown at five captions separated clearly from that of the matrix prediction as single-caption reliability decreased and the sample size increased; for example, 0.0146 versus 0.0077 for $n = 2{,}000$ and $A_y(1) = 0.3$. At $A_y(1) = 0.8$ the two predictions gave the same error. The difference predicted by Proposition~\ref{prop:sb} thus also appears in finite samples.

\paragraph{Prediction on COCO (K1).} The prediction built from two captions matched the values observed on the separate image half with the mean of three, four and five captions with mean absolute errors of 0.009, 0.014 and 0.014. At $r = 4$ and $r = 5$, however, the upper end of the interval exceeded the tolerance of 0.03, and the test was not passed. The signed errors of the two prediction rules have opposite signs: the matrix prediction lay on average 0.009--0.014 above the observed values and the Spearman--Brown prediction 0.009--0.014 below them. As explained in Section~\ref{sec:prediction}, the form of the prediction holds exactly at the population level under exchangeability; what K1 tests is precision in finite samples. With 1,600 images in each half and with prediction and observation made on different images, the sampling difference between the two halves adds to the error and widens the intervals. As the plan requires, $\CKAs$ on COCO is interpreted not as the limit of averaging but as a model-based estimate of the alignment between the shared components of the captions. Building on this experience, the M1 test on MSR-VTT was designed as a paired test.

\begin{figure}[t]
\centering
\includegraphics[width=\linewidth]{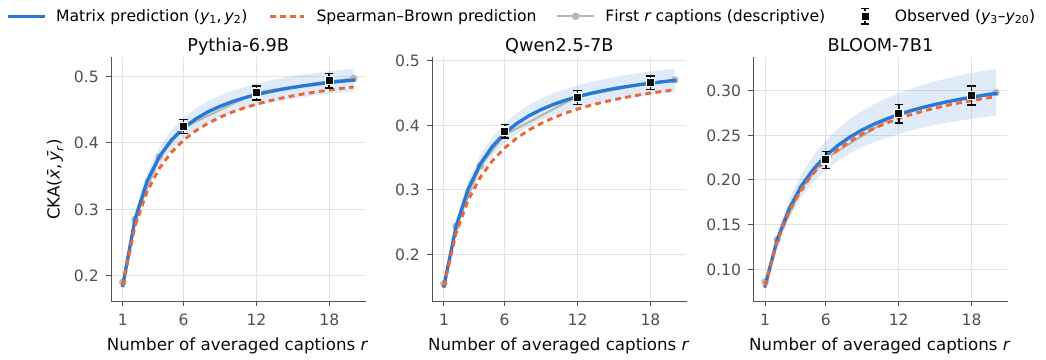}
\caption{Averaging curves on MSR-VTT (DINOv2-g and the three 7B language models, 5,000 clips). The blue line and band are the matrix prediction built from the first two captions only and its 95\% interval; the orange dashed line is the Spearman--Brown prediction computed from the same components. Black squares are the values observed on the same clips with the mean of 6, 12 and 18 captions drawn at random from the remaining 18. The grey line shows, descriptively, the values observed with the mean of the first $r$ captions.}
\label{fig:m1}
\end{figure}

\paragraph{Prediction on MSR-VTT (M1).} The prediction built from only two captions matched the alignment observed on the same clips with the mean of 6, 12 and 18 captions drawn from the remaining captions with mean absolute errors of 0.003, 0.004 and 0.004 (Figure~\ref{fig:m1}). The upper ends of the intervals were 0.010, 0.013 and 0.014, all clearly below the tolerance of 0.03; M1 was passed. The signed error of the matrix prediction is close to zero at all three values of $r$: $+0.0002$ [$-0.009$, 0.009], $+0.0003$ [$-0.011$, 0.011] and $+0.001$ [$-0.011$, 0.012]. The point estimate of the mean absolute error lies at the lower end of its bootstrap interval. This is because the absolute value accumulates sampling noise in the positive direction across resampled data, which shifts the interval upward and makes the decision rule conservative.

The Spearman--Brown prediction, by contrast, fell systematically below the observed values. Its signed errors were $+0.012$ [0.004, 0.020], $+0.010$ [0.0002, 0.020] and $+0.009$ [$-0.002$, 0.019], and the interval excluded zero at $r = 6$ and $r = 12$. The direction of the deviation is the one predicted by Proposition~\ref{prop:sb}, and it is the same on COCO: in both datasets the Spearman--Brown prediction lay below the matrix prediction. The M1 result supports two conclusions: the improvement of the matrix form over the classical correction is of measurable size, and the failure of K1 stems from limited precision rather than from the method.

\subsection{Observation reliability and model comparisons}
\label{sec:reliability}

\begin{figure}[t]
\centering
\includegraphics[width=\linewidth]{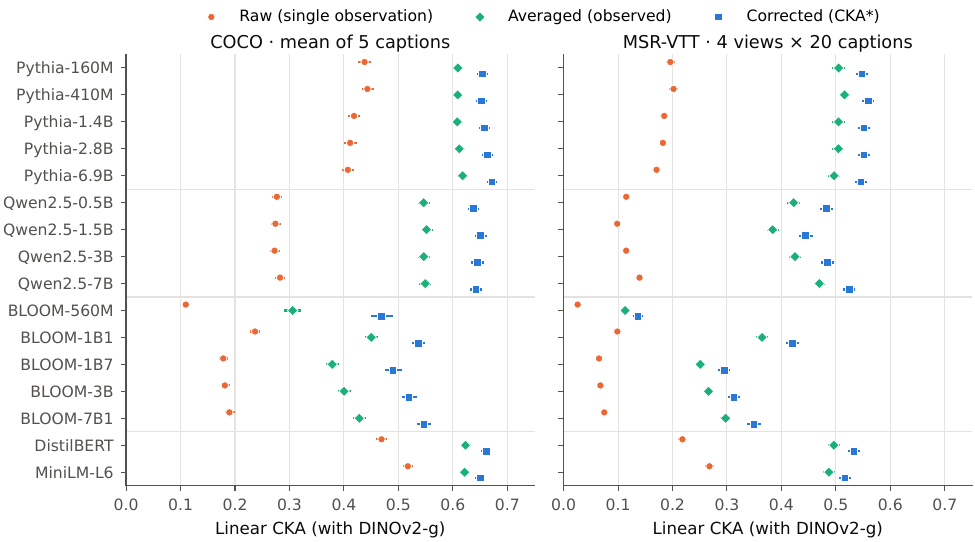}
\caption{Raw, averaged and corrected alignment of DINOv2-g with each language model. Horizontal bars are 95\% bootstrap intervals. Averaging and correction remove much of the spread in raw measurement; the raw advantage of MiniLM-L6 disappears under corrected measurement.}
\label{fig:raw}
\end{figure}

Table~\ref{tab:models} and Figure~\ref{fig:raw} give, for both datasets, the raw, averaged and corrected alignment of DINOv2-g with each language model, together with the single-caption reliability factor. The half-widths of the 95\% intervals of the alignment values are at most 0.020 on COCO and 0.013 on MSR-VTT.

\begin{table}[t]
\centering
\caption{Alignment of DINOv2-g with each language model. Raw: CKA with single observations (on MSR-VTT, the mean over four view--caption pairs). Avg.: averaged CKA (COCO: five captions; MSR-VTT: four views and 20 captions). $A_y(1)$: single-caption reliability factor. $\CKAs$: corrected alignment. Layer: selected layer / total layers. Params: parameter count excluding the language-modeling head (billions).}
\label{tab:models}
\small
\setlength{\tabcolsep}{4.2pt}
\begin{tabular}{@{}l r c cccc cccc@{}}
\toprule
 & & & \multicolumn{4}{c}{COCO (3,200 images)} & \multicolumn{4}{c}{MSR-VTT (5,000 clips)} \\
\cmidrule(lr){4-7} \cmidrule(l){8-11}
Model & Params & Layer & Raw & Avg. & $A_y(1)$ & $\CKAs$ & Raw & Avg. & $A_y(1)$ & $\CKAs$ \\
\midrule
Pythia-160M & 0.12 & 12/12 & 0.438 & 0.609 & 0.669 & 0.655 & 0.196 & 0.505 & 0.399 & 0.548 \\
Pythia-410M & 0.35 & 24/24 & 0.443 & 0.609 & 0.679 & 0.652 & 0.202 & 0.516 & 0.405 & 0.560 \\
Pythia-1.4B & 1.31 & 24/24 & 0.419 & 0.608 & 0.636 & 0.658 & 0.185 & 0.505 & 0.375 & 0.552 \\
Pythia-2.8B & 2.65 & 32/32 & 0.411 & 0.612 & 0.620 & 0.663 & 0.182 & 0.505 & 0.370 & 0.552 \\
Pythia-6.9B & 6.65 & 32/32 & 0.407 & 0.618 & 0.606 & 0.672 & 0.171 & 0.497 & 0.350 & 0.546 \\
\addlinespace
Qwen2.5-0.5B & 0.49 & 24/24 & 0.277 & 0.546 & 0.434 & 0.638 & 0.115 & 0.423 & 0.265 & 0.483 \\
Qwen2.5-1.5B & 1.54 & 28/28 & 0.274 & 0.552 & 0.421 & 0.651 & 0.098 & 0.384 & 0.246 & 0.444 \\
Qwen2.5-3B & 3.09 & 36/36 & 0.273 & 0.547 & 0.422 & 0.645 & 0.115 & 0.425 & 0.263 & 0.485 \\
Qwen2.5-7B & 7.07 & 28/28 & 0.283 & 0.549 & 0.439 & 0.642 & 0.139 & 0.470 & 0.295 & 0.524 \\
\addlinespace
BLOOM-560M & 0.56 & 24/24 & 0.110 & 0.306 & 0.234 & 0.469 & 0.026 & 0.113 & 0.204 & 0.137 \\
BLOOM-1B1 & 1.07 & 24/24 & 0.237 & 0.450 & 0.441 & 0.537 & 0.099 & 0.364 & 0.262 & 0.420 \\
BLOOM-1B7 & 1.72 & 24/24 & 0.178 & 0.379 & 0.364 & 0.490 & 0.065 & 0.251 & 0.248 & 0.295 \\
BLOOM-3B & 3.00 & 30/30 & 0.181 & 0.400 & 0.348 & 0.520 & 0.068 & 0.266 & 0.244 & 0.313 \\
BLOOM-7B1 & 7.07 & 30/30 & 0.190 & 0.428 & 0.346 & 0.547 & 0.075 & 0.298 & 0.240 & 0.350 \\
\addlinespace
DistilBERT & 0.07 & 5/6 & 0.469 & 0.623 & 0.709 & 0.661 & 0.218 & 0.496 & 0.456 & 0.533 \\
MiniLM-L6 & 0.02 & 6/6 & 0.517 & 0.622 & 0.796 & 0.650 & 0.268 & 0.487 & 0.577 & 0.516 \\
\bottomrule
\end{tabular}
\end{table}

\paragraph{Reliability loss is concentrated on the text side.} The single-caption reliability factor ranged from 0.234 to 0.796 on COCO and from 0.204 to 0.577 on MSR-VTT. The ordering at the family level is the same in both datasets: reliability is highest for MiniLM-L6, followed by DistilBERT and the Pythia family, and lowest for the Qwen2.5 and BLOOM families. The video side is far more reliable: on MSR-VTT, the reliability factor of a single view ranged from 0.819 to 0.907 across encoders, and that of the mean of four views from 0.949 to 0.976. Almost all of the loss in raw alignment therefore comes from the text side.

\paragraph{Averaging and correction narrow the differences.} For the eleven models outside BLOOM, raw alignment on COCO spans 0.273--0.517, narrowing to 0.546--0.623 with the mean of five captions and to 0.638--0.672 after correction. On MSR-VTT, the corresponding ranges for the same models are 0.098--0.268, 0.384--0.516 and 0.444--0.560. The BLOOM family remains lower on all three measures. The Spearman rank correlation between the raw and corrected rankings was 0.80 [0.73, 0.85] on COCO and 0.86 [0.85, 0.88] on MSR-VTT. The two rankings are related but not interchangeable.

\paragraph{Ranking reversal (M2a).} The raw alignment of MiniLM-L6 exceeded the highest of the three 7B models by 0.097 [0.092, 0.103], replicating the COCO finding. This advantage is specific to raw measurement. In corrected alignment, MiniLM-L6 (0.516) fell behind Pythia-6.9B (0.546) and Qwen2.5-7B (0.524) on MSR-VTT, and on COCO it lay just below the Pythia models (0.650 versus 0.652--0.672). The source of the advantage is reliability. MiniLM-L6 gives far more consistent representations to different captions of the same clip: the single-caption reliability factor is 0.577 for MiniLM-L6 and 0.350 for Pythia-6.9B.

\paragraph{Family differences (M2b).} The difference between the Pythia and Qwen2.5 family means was 0.070 [0.067, 0.074] under raw measurement and 0.080 [0.075, 0.085] under averaged measurement. Instead of shrinking, the difference grew by 0.010 [0.005, 0.015], and the prediction was reversed. On COCO, the corrected alignments of the two families were very close, and most of the raw difference came from the difference in reliability. On MSR-VTT, by contrast, the corrected alignments also separate clearly (Pythia 0.546--0.560; Qwen2.5 0.444--0.524). As a measure that was not prespecified, the relative difference did shrink: the ratio of the Pythia mean to the Qwen2.5 mean is about 1.60 under raw and 1.19 under averaged measurement. Because the decomposition is multiplicative, rising reliability scales all values upward, and an absolute difference can grow even as the relative difference shrinks. Defining M2b on the absolute difference was not consistent with this structure.

\subsection{Decomposing scale trends}
\label{sec:scale}

\paragraph{COCO (H2).} The raw slope was positive for BLOOM, 0.041 [0.035, 0.047], and for Qwen2.5, 0.004 [0.0001, 0.007], and negative for Pythia, $-0.022$ [$-0.024$, $-0.019$]; the precondition was met in two families. The decomposition revealed a different picture in each family. In BLOOM both components increased with scale: the slope of $\log A_y(1)$ was 0.202 [0.163, 0.239] and that of $\log \CKAs$ 0.108 [0.082, 0.137]. About two thirds of the total increase on the log scale comes from reliability. In Qwen2.5 the raw slope is very small, and the slope intervals of both components contain zero. In Pythia, raw alignment decreased with scale because reliability decreased (slope of $\log A_y(1)$: $-0.066$ [$-0.071$, $-0.062$]). Corrected alignment, by contrast, increased slightly (slope of $\log \CKAs$: 0.015 [0.011, 0.019]). The directly observed five-caption averages support this direction: 0.608--0.609 for the three smallest models and 0.618 for Pythia-6.9B.

\paragraph{MSR-VTT (M2c).} The decline of raw alignment with scale in the Pythia family replicated: slope $-0.016$ [$-0.018$, $-0.014$]. Corrected alignment, however, did not increase as on COCO but stayed flat between 0.546 and 0.560 (slope of $\log \CKAs$: $-0.005$ [$-0.009$, $-0.0001$]). As the plan prescribes, the outcome is partial.

\begin{figure}[t]
\centering
\includegraphics[width=\linewidth]{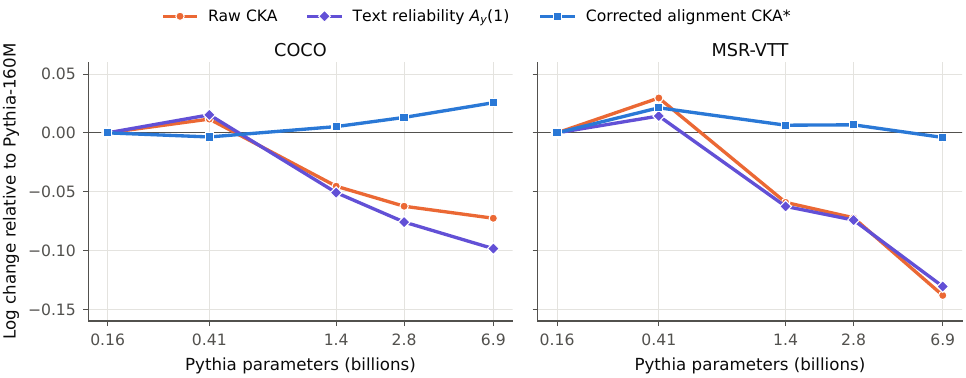}
\caption{Decomposition of the scale trend in the Pythia family (DINOv2-g). Each quantity is shown as the log change relative to its value for Pythia-160M. Because the decomposition is additive on the log scale, the change in raw CKA is approximately the sum of the changes in text reliability and corrected alignment; the reliability of the video side is the same for all language models. In both datasets, the raw decline moves together with the decline in reliability. Corrected alignment increases slightly on COCO and stays flat on MSR-VTT. Slope intervals are given in the text.}
\label{fig:pythia}
\end{figure}

\paragraph{The finding common to both datasets.} Figure~\ref{fig:pythia} shows the decomposition as log changes relative to Pythia-160M. From Pythia-160M to Pythia-6.9B, the logarithm of raw alignment fell by 0.073 on COCO and by 0.138 on MSR-VTT. Over the same range, the logarithm of text reliability fell by 0.098 on COCO and by 0.131 on MSR-VTT. In both datasets, the decline in raw alignment is accounted for by the decline in text reliability. The change of corrected alignment with scale is small, and its direction depends on the dataset: the log change is $+0.026$ on COCO and $-0.004$ on MSR-VTT.

\subsection{The residual gap and the within-modality reference}
\label{sec:gap}

\paragraph{COCO (H1).} The highest within-modality alignment of DINOv2-g with an encoder from a different family was 0.679 [0.670, 0.688], obtained with ViT-L/16 (ImageNet-21k). The MAE models aligned with DINOv2-g only at about 0.12. The mean gap of the three 7B language models was $G = 0.058$ [0.045, 0.072]. Since the interval lies neither entirely above $\delta = 0.05$ nor entirely within $\pm\delta$, H1 is inconclusive. The gap varies widely across language models: 0.007 [$-0.006$, 0.020] for Pythia-6.9B, 0.036 [0.023, 0.050] for Qwen2.5-7B and 0.132 [0.116, 0.147] for BLOOM-7B1. The corrected alignment of Pythia-6.9B lies within the equivalence margin of the alignment of DINOv2-g with the strongest image encoder from a different family. The positive mean gap is driven largely by BLOOM-7B1.

\paragraph{MSR-VTT (exploratory).} When video encoders are compared under the disjoint-observation rule, the highest corrected alignment of DINOv2-g with an encoder from a different family was 0.684 [0.678, 0.691], obtained with ViT-B/16 (ImageNet-21k). Alignment within the DINOv2 family was about 0.89--0.94, whereas the MAE models aligned with DINOv2-g only at about 0.14. Against this reference, the mean gap of the three 7B language models was $G = 0.211$ [0.201, 0.221]; the gap was 0.139 for Pythia-6.9B, 0.160 for Qwen2.5-7B and 0.335 for BLOOM-7B1. Even after observation noise is corrected, video--text alignment remains below the highest alignment between video encoders from different families.

\paragraph{Effect of the disjoint-observation rule (exploratory).} How within-modality alignment is computed changes the result substantially. On COCO, across the 120 pairs of language models, the median raw CKA computed on the same caption was 0.574 (interquartile range 0.449--0.672), and that computed on different captions of the same image 0.193 (0.138--0.262). The median corrected alignment under the disjoint-observation rule was 0.865 (0.799--0.928), and 0.951 for the 26 pairs within the three decoder families. On MSR-VTT, the same three values were 0.525 (0.381--0.633), 0.070 (0.052--0.098) and 0.763 (0.647--0.859), and 0.926 within families. Raw values on the same caption exceed those on different captions by a factor of about three on COCO and of more than seven on MSR-VTT. Given an effect of shared observations of this size, a within-modality alignment computed on the same observation cannot be compared on an equal footing with a cross-modal alignment that never shares observations.

\subsection{Choices of metric, layer and pooling}
\label{sec:choices}

The analyses in this section were not prespecified and do not enter the decision rules.

\begin{figure}[t]
\centering
\includegraphics[width=\linewidth]{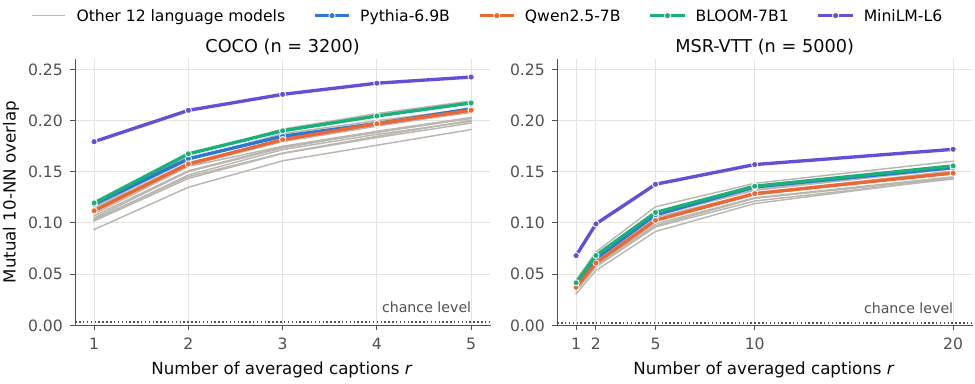}
\caption{Mutual 10-nearest-neighbor overlap between DINOv2-g and the language models as a function of the number of averaged captions. Colored lines show the three 7B models and MiniLM-L6; grey lines show the other twelve models. The dotted line is the overlap expected under random matching.}
\label{fig:knn}
\end{figure}

\paragraph{Mutual nearest neighbors.} The effect of observation noise is not specific to linear CKA. The mutual 10-nearest-neighbor overlap used by Huh et al.\ also increased as captions were averaged, in both datasets and for all 16 language models (Figure~\ref{fig:knn}). For Pythia-6.9B, for example, the overlap rose from 0.118 with a single caption to 0.211 with the mean of five captions on COCO, and from 0.038 to 0.154 with the mean of 20 captions on MSR-VTT. The raw advantage of MiniLM-L6 was also visible under this metric, and its distance to the other models narrowed as captions were averaged. Differences between language models are markedly smaller under this metric than under CKA: although BLOOM-560M has the lowest corrected CKA on MSR-VTT (0.137), its neighborhood overlap with 20 captions (0.149) is close to that of the other models.

\paragraph{Layer selection.} On COCO, selecting layers by corrected alignment instead of raw CKA would have changed the selected layer for nine of the 16 models. For the BLOOM models, layers three or four before the last gave higher corrected alignment on the selection images; for BLOOM-7B1, for example, 0.650 at layer 26 versus 0.556 at layer 30. The low corrected alignment of BLOOM in Table~\ref{tab:models} therefore depends partly on the layer rule. In some intermediate layers, on the other hand, raw alignment was very close to zero while the corrected value was very high: at layer 5 of Pythia-160M, raw CKA is 0.004 and $\CKAs$ is 0.919, which corresponds to a reliability factor below 0.01. When the shared variance between captions is this small, the corrected ratio rests on a denominator close to zero and should not be interpreted. The smallest reliability factor in the main analysis is 0.234 on COCO and 0.204 on MSR-VTT.

\paragraph{Pooling.} When the class token of DINOv2-g was used instead of the patch mean, all values decreased, but the ranking of the three 7B models did not change. Raw and corrected alignment on COCO were 0.274 and 0.451 for Pythia-6.9B, 0.167 and 0.381 for Qwen2.5-7B, and 0.109 and 0.315 for BLOOM-7B1; on MSR-VTT they were 0.161 and 0.502, 0.128 and 0.472, and 0.071 and 0.322, respectively.

\section{Discussion}

\paragraph{Raw alignment measures two properties at once.} The most consistent aspect of the findings is that alignment measured with single observations conflates two distinct properties: how well the shared component of the scene agrees across the two modalities, and how consistently the language model represents different captions of the same scene. The second property varies greatly across models. While the image and video side is far more reliable, the single-caption reliability factor varies by a factor of about three across language models. The direction of the differences is also meaningful. MiniLM-L6, a sentence encoder trained with a contrastive objective to map semantically similar sentences to nearby representations, showed the highest reliability. A raw comparison of the form ``this language model aligns better with images'' therefore partly means ``this model's representation is less sensitive to the wording of the caption''. Corrected alignment separates the two properties and measures the first on its own.

\paragraph{Scale trends and the Platonic representation hypothesis.} \citet{huh2024platonic} reported that the alignment of language models with images increases with model capability, whereas \citet{koepke2026cave} found that this trend does not appear for newer models. Our findings point to a possible source of such discrepancies. On the log scale, the change of raw alignment with scale is the sum of the changes in reliability and in corrected alignment, and these two components behave differently across families. In BLOOM, about two thirds of the raw increase comes from reliability, whereas in Pythia the decline in reliability pulled raw alignment down and masked a small change in corrected alignment. A raw scale trend should therefore be split into these two components before it is read as evidence that representations converge or diverge. \citet{balestriero2026proving} used two independent observations of the same scene, for example two captions of the same image, as good views. This is the same structure as the repeated observations used here for measurement; the difference is that it serves training there and evaluation here. There, the fraction explained by the scene is a property of the representation the encoder has learned and bounds the attainable alignment. The reliability measured here, in contrast, is an effect that lowers the raw measurement at evaluation time and can be corrected. The two views complement each other: one concerns the scene information an encoder represents, the other how reliably this information is measured with a single observation.

\paragraph{Averaging and many-to-many matching.} \citet{he2025seeing} observed that alignment increases when exemplars are averaged, whereas \citet{koepke2026cave} reported that mutual nearest-neighbor alignment drops in the regime with several captions per image. The measurement model explains both findings. When each caption is matched as a separate item, the multiple valid captions of the same image blur the neighborhood structure. When captions are treated as repeated observations of the same scene and averaged, observation noise is reduced and alignment increases. Our analysis also predicts the size of the increase: the gain from averaging is determined by the scene and noise covariances estimated from two captions, and on MSR-VTT this prediction was confirmed against direct observation with a mean error of at most 0.004. The mutual nearest-neighbor measure also increased consistently with averaging in both datasets.

\paragraph{The residual gap depends on the reference.} Even after observation noise is corrected, cross-modal alignment is not perfect, but the size of the gap depends both on the language model and on the within-modality pair taken as the reference. On COCO, Pythia-6.9B came within the equivalence margin of the alignment of DINOv2-g with the strongest image encoder from a different family, whereas BLOOM-7B1 remained clearly behind in both datasets. The within-modality reference is not a single value either: the MAE models align with DINOv2-g only at 0.12--0.14, and encoders of the same modality from different families are far from converging. Moreover, the reference must be computed on disjoint observations; otherwise, shared observations artificially inflate the within-modality value. The answer to the question whether cross-modal alignment reaches the within-modality level therefore depends on which reference and which measurement rule are chosen.

\paragraph{Practical recommendations.} The findings yield four practical recommendations for studies that measure cross-modal alignment. First, at least two observations per scene should be used, and the reliability factor of each modality should be reported alongside the alignment. Second, the averaging curve should be predicted with the matrix form, which accounts for the general shape of the noise, rather than with Spearman--Brown; Proposition~\ref{prop:sb} and the M1 results show that the classical formula systematically understates the gain. Third, the disjoint-observation rule should be applied when two models of the same modality are compared. Fourth, the bootstrap for U-statistic estimators should use multiplicity weights rather than duplicated rows.

\paragraph{Limitations.} The results should be read with several limitations in mind. The analyses use linear CKA, which is sensitive to high-variance directions and, as the BLOOM-560M example shows, does not always yield the same ranking as neighborhood-based measures. Using the captions in random order supports A1. However, since shared tendencies of annotators are absorbed into the scene component, $\CKAs$ is relative to a particular captioning protocol, and the COCO and MSR-VTT protocols differ. On MSR-VTT, A2, the uncorrelatedness of caption noise and frame-selection noise, is reasonable but has not been tested directly. The study covers only frozen pretrained models and English captions. Language-model layers were selected by raw CKA, and as the BLOOM example shows, this choice can affect the corrected values. The M2b prediction was defined on the absolute difference; given the multiplicative decomposition, a logarithmic or ratio-based definition would have been more appropriate. Finally, the analysis plans were locked locally. Although the locking mechanism and all hashes are released, the freezing of the plans before the results is not documented by an independent timestamp.

\section{Conclusion}

This paper has examined the role of observation noise in measuring cross-modal representational alignment. The measurement model, based on repeated observations of the same scene, factorizes observed linear CKA into corrected alignment and the reliability factors of the two modalities. The general form of the reliability factor contains the Spearman--Brown correction as a special case, and we prove that this correction understates the gain from averaging whenever the noise is not proportional to the scene component. In two studies with analysis plans locked in advance, the prediction built from two captions matched the alignment observed directly on MSR-VTT, whereas Spearman--Brown deviated in the direction predicted by theory. Reliability differences changed the ranking of language models and the interpretation of scale trends; some findings replicated across the two datasets, others depended on the dataset. Testing the Platonic representation hypothesis requires attention not only to the representations themselves but also to the reliability of the data through which they are observed. The proposed decomposition and estimators offer a tool for making this distinction a routine part of model comparisons.

\section*{Data and code availability}

COCO and MSR-VTT are publicly available datasets. Model weights were obtained from pinned revisions on Hugging Face. The analysis notebooks, the locked analysis plans with their SHA-256 hashes, the pilot and full-stage outputs, and scripts that verify the hash chain and reproduce the tables and figures of this paper from the released outputs are available at \url{https://github.com/fatihkutlu/lossy-platonic-alignment}. The version used in this paper (v1.0.0) is archived on Zenodo at \url{https://doi.org/10.5281/zenodo.24482107}.

\section*{Declaration of generative AI use}

During the preparation of this work, the author used Claude (Anthropic) in order to draft and revise the text, improve its language and readability, and translate the manuscript from Turkish into English. After using this tool, the author reviewed and edited the content as needed and takes full responsibility for the content of the article.

\bibliography{references}

\begin{thebibliography}{43}
\providecommand{\natexlab}[1]{#1}
\providecommand{\url}[1]{\texttt{#1}}
\expandafter\ifx\csname urlstyle\endcsname\relax
  \providecommand{\doi}[1]{doi: #1}\else
  \providecommand{\doi}{doi: \begingroup \urlstyle{rm}\Url}\fi

\bibitem[Balestriero and LeCun(2026)]{balestriero2026proving}
Randall Balestriero and Yann LeCun.
\newblock Proving and improving the {P}latonic representation hypothesis with
  {JEPA}s.
\newblock arXiv:2610.39302, 2026.

\bibitem[Biderman et~al.(2023)Biderman, Schoelkopf, Anthony, Bradley, O'Brien,
  Hallahan, Khan, Purohit, Prashanth, Raff, Skowron, Sutawika, and van~der
  Wal]{biderman2023pythia}
Stella Biderman, Hailey Schoelkopf, Quentin Anthony, Herbie Bradley, Kyle
  O'Brien, Eric Hallahan, Mohammad~Aflah Khan, Shivanshu Purohit, USVSN~Sai
  Prashanth, Edward Raff, Aviya Skowron, Lintang Sutawika, and Oskar van~der
  Wal.
\newblock Pythia: A suite for analyzing large language models across training
  and scaling.
\newblock In \emph{Proceedings of the 40th International Conference on Machine
  Learning}, volume 202 of \emph{Proceedings of Machine Learning Research},
  pages 2397--2430, 2023.

\bibitem[{BigScience Workshop}(2024)]{bigscience2024bloom}
{BigScience Workshop}.
\newblock {BLOOM}: A 176{B}-parameter open-access multilingual language model.
\newblock \emph{Journal of Machine Learning Research}, 25\penalty0
  (422):\penalty0 1--74, 2024.

\bibitem[Brown(1910)]{brown1910some}
William Brown.
\newblock Some experimental results in the correlation of mental abilities.
\newblock \emph{British Journal of Psychology}, 3\penalty0 (3):\penalty0
  296--322, 1910.

\bibitem[Chen et~al.(2015)Chen, Fang, Lin, Vedantam, Gupta, Doll{\'a}r, and
  Zitnick]{chen2015coco}
Xinlei Chen, Hao Fang, Tsung-Yi Lin, Ramakrishna Vedantam, Saurabh Gupta, Piotr
  Doll{\'a}r, and C.~Lawrence Zitnick.
\newblock Microsoft {COCO} captions: Data collection and evaluation server.
\newblock arXiv:1504.00325, 2015.

\bibitem[Chun et~al.(2025)Chun, Canatar, Chung, and Lee]{chun2025estimating}
Chanwoo Chun, Abdulkadir Canatar, SueYeon Chung, and Daniel~D. Lee.
\newblock Estimating neural representation alignment from sparsely sampled
  inputs and features.
\newblock arXiv:2502.15104, 2025.

\bibitem[Cortes et~al.(2012)Cortes, Mohri, and
  Rostamizadeh]{cortes2012algorithms}
Corinna Cortes, Mehryar Mohri, and Afshin Rostamizadeh.
\newblock Algorithms for learning kernels based on centered alignment.
\newblock \emph{Journal of Machine Learning Research}, 13:\penalty0 795--828,
  2012.

\bibitem[Davari et~al.(2023)Davari, Horoi, Natik, Lajoie, Wolf, and
  Belilovsky]{davari2023reliability}
MohammadReza Davari, Stefan Horoi, Amine Natik, Guillaume Lajoie, Guy Wolf, and
  Eugene Belilovsky.
\newblock Reliability of {CKA} as a similarity measure in deep learning.
\newblock In \emph{International Conference on Learning Representations}, 2023.

\bibitem[Ding et~al.(2021)Ding, Denain, and Steinhardt]{ding2021grounding}
Frances Ding, Jean-Stanislas Denain, and Jacob Steinhardt.
\newblock Grounding representation similarity through statistical testing.
\newblock In \emph{Advances in Neural Information Processing Systems},
  volume~34, 2021.

\bibitem[Dosovitskiy et~al.(2021)Dosovitskiy, Beyer, Kolesnikov, Weissenborn,
  Zhai, Unterthiner, Dehghani, Minderer, Heigold, Gelly, Uszkoreit, and
  Houlsby]{dosovitskiy2021image}
Alexey Dosovitskiy, Lucas Beyer, Alexander Kolesnikov, Dirk Weissenborn,
  Xiaohua Zhai, Thomas Unterthiner, Mostafa Dehghani, Matthias Minderer, Georg
  Heigold, Sylvain Gelly, Jakob Uszkoreit, and Neil Houlsby.
\newblock An image is worth 16x16 words: Transformers for image recognition at
  scale.
\newblock In \emph{International Conference on Learning Representations}, 2021.

\bibitem[Efron and Tibshirani(1993)]{efron1993introduction}
Bradley Efron and Robert~J. Tibshirani.
\newblock \emph{An Introduction to the Bootstrap}.
\newblock Chapman \& Hall, New York, 1993.

\bibitem[Gretton et~al.(2005)Gretton, Bousquet, Smola, and
  Sch{\"o}lkopf]{gretton2005measuring}
Arthur Gretton, Olivier Bousquet, Alex Smola, and Bernhard Sch{\"o}lkopf.
\newblock Measuring statistical dependence with {H}ilbert--{S}chmidt norms.
\newblock In \emph{Algorithmic Learning Theory}, volume 3734 of \emph{Lecture
  Notes in Computer Science}, pages 63--77. Springer, 2005.

\bibitem[Gr{\"o}ger et~al.(2026)Gr{\"o}ger, Wen, and
  Brbi{\'c}]{groger2026aristotelian}
Fabian Gr{\"o}ger, Shuo Wen, and Maria Brbi{\'c}.
\newblock Revisiting the {P}latonic representation hypothesis: An
  {A}ristotelian view.
\newblock In \emph{Proceedings of the 43rd International Conference on Machine
  Learning}, volume 306 of \emph{Proceedings of Machine Learning Research},
  pages 36724--36766, 2026.

\bibitem[He et~al.(2022)He, Chen, Xie, Li, Doll{\'a}r, and
  Girshick]{he2022masked}
Kaiming He, Xinlei Chen, Saining Xie, Yanghao Li, Piotr Doll{\'a}r, and Ross
  Girshick.
\newblock Masked autoencoders are scalable vision learners.
\newblock In \emph{Proceedings of the IEEE/CVF Conference on Computer Vision
  and Pattern Recognition}, pages 16000--16009, 2022.

\bibitem[He et~al.(2025)He, Trott, and Khosla]{he2025seeing}
Zoe~Wanying He, Sean Trott, and Meenakshi Khosla.
\newblock Seeing through words, speaking through pixels: Deep representational
  alignment between vision and language models.
\newblock In \emph{Proceedings of the 2025 Conference on Empirical Methods in
  Natural Language Processing}, pages 35657--35672, 2025.

\bibitem[Huh et~al.(2024)Huh, Cheung, Wang, and Isola]{huh2024platonic}
Minyoung Huh, Brian Cheung, Tongzhou Wang, and Phillip Isola.
\newblock Position: The {P}latonic representation hypothesis.
\newblock In \emph{Proceedings of the 41st International Conference on Machine
  Learning}, volume 235 of \emph{Proceedings of Machine Learning Research},
  2024.

\bibitem[Kang et~al.(2025)Kang, Canatar, and Chung]{kang2025spectral}
Hyunmo Kang, Abdulkadir Canatar, and SueYeon Chung.
\newblock Spectral analysis of representational similarity with limited
  neurons.
\newblock In \emph{Advances in Neural Information Processing Systems},
  volume~38, 2025.

\bibitem[Klabunde et~al.(2025)Klabunde, Schumacher, Strohmaier, and
  Lemmerich]{klabunde2025similarity}
Max Klabunde, Tobias Schumacher, Markus Strohmaier, and Florian Lemmerich.
\newblock Similarity of neural network models: A survey of functional and
  representational measures.
\newblock \emph{ACM Computing Surveys}, 57\penalty0 (9), 2025.

\bibitem[Koepke et~al.(2026)Koepke, Zverev, Ginosar, and Efros]{koepke2026cave}
A.~Sophia Koepke, Daniil Zverev, Shiry Ginosar, and Alexei~A. Efros.
\newblock Back into {P}lato's cave: Examining cross-modal representational
  convergence at scale.
\newblock arXiv:2604.18572, 2026.

\bibitem[Kornblith et~al.(2019)Kornblith, Norouzi, Lee, and
  Hinton]{kornblith2019similarity}
Simon Kornblith, Mohammad Norouzi, Honglak Lee, and Geoffrey Hinton.
\newblock Similarity of neural network representations revisited.
\newblock In \emph{Proceedings of the 36th International Conference on Machine
  Learning}, volume~97 of \emph{Proceedings of Machine Learning Research},
  pages 3519--3529, 2019.

\bibitem[Kriegeskorte et~al.(2008)Kriegeskorte, Mur, and
  Bandettini]{kriegeskorte2008rsa}
Nikolaus Kriegeskorte, Marieke Mur, and Peter Bandettini.
\newblock Representational similarity analysis -- connecting the branches of
  systems neuroscience.
\newblock \emph{Frontiers in Systems Neuroscience}, 2:\penalty0 4, 2008.

\bibitem[Lage-Castellanos et~al.(2019)Lage-Castellanos, Valente, Formisano, and
  De~Martino]{lage2019methods}
Agustin Lage-Castellanos, Giancarlo Valente, Elia Formisano, and Federico
  De~Martino.
\newblock Methods for computing the maximum performance of computational models
  of f{MRI} responses.
\newblock \emph{PLoS Computational Biology}, 15\penalty0 (3):\penalty0
  e1006397, 2019.

\bibitem[Lin et~al.(2014)Lin, Maire, Belongie, Hays, Perona, Ramanan,
  Doll{\'a}r, and Zitnick]{lin2014coco}
Tsung-Yi Lin, Michael Maire, Serge Belongie, James Hays, Pietro Perona, Deva
  Ramanan, Piotr Doll{\'a}r, and C.~Lawrence Zitnick.
\newblock Microsoft {COCO}: Common objects in context.
\newblock In \emph{European Conference on Computer Vision}, pages 740--755,
  2014.

\bibitem[Lord and Novick(1968)]{lord1968statistical}
Frederic~M. Lord and Melvin~R. Novick.
\newblock \emph{Statistical Theories of Mental Test Scores}.
\newblock Addison-Wesley, Reading, MA, 1968.

\bibitem[Maniparambil et~al.(2024)Maniparambil, Akshulakov, Djilali, Seddik,
  Narayan, Mangalam, and O'Connor]{maniparambil2024vision}
Mayug Maniparambil, Raiymbek Akshulakov, Yasser Abdelaziz~Dahou Djilali,
  Mohamed El~Amine Seddik, Sanath Narayan, Karttikeya Mangalam, and Noel~E.
  O'Connor.
\newblock Do vision and language encoders represent the world similarly?
\newblock In \emph{Proceedings of the IEEE/CVF Conference on Computer Vision
  and Pattern Recognition}, 2024.

\bibitem[Muzellec et~al.(2026)Muzellec, Alghetaa, Kornblith, and
  Kar]{muzellec2026masked}
Sabine Muzellec, Yousif~Kashef Alghetaa, Simon Kornblith, and Kohitij Kar.
\newblock Masked attribution-based probing of strategies as a computational
  framework to align human, non-human primate, and model explanations.
\newblock \emph{Communications Psychology}, 4:\penalty0 158, 2026.

\bibitem[Nguyen et~al.(2021)Nguyen, Raghu, and Kornblith]{nguyen2021wide}
Thao Nguyen, Maithra Raghu, and Simon Kornblith.
\newblock Do wide and deep networks learn the same things? uncovering how
  neural network representations vary with width and depth.
\newblock In \emph{International Conference on Learning Representations}, 2021.

\bibitem[Nili et~al.(2014)Nili, Wingfield, Walther, Su, Marslen-Wilson, and
  Kriegeskorte]{nili2014toolbox}
Hamed Nili, Cai Wingfield, Alexander Walther, Li~Su, William Marslen-Wilson,
  and Nikolaus Kriegeskorte.
\newblock A toolbox for representational similarity analysis.
\newblock \emph{PLoS Computational Biology}, 10\penalty0 (4):\penalty0
  e1003553, 2014.

\bibitem[Nosek et~al.(2018)Nosek, Ebersole, DeHaven, and
  Mellor]{nosek2018preregistration}
Brian~A. Nosek, Charles~R. Ebersole, Alexander~C. DeHaven, and David~T. Mellor.
\newblock The preregistration revolution.
\newblock \emph{Proceedings of the National Academy of Sciences}, 115\penalty0
  (11):\penalty0 2600--2606, 2018.

\bibitem[Oquab et~al.(2024)]{oquab2024dinov2}
Maxime Oquab et~al.
\newblock {DINOv2}: Learning robust visual features without supervision.
\newblock \emph{Transactions on Machine Learning Research}, 2024.

\bibitem[{Qwen Team}(2024)]{qwen2024qwen25}
{Qwen Team}.
\newblock Qwen2.5 technical report.
\newblock arXiv:2412.15115, 2024.

\bibitem[Radford et~al.(2021)Radford, Kim, Hallacy, Ramesh, Goh, Agarwal,
  Sastry, Askell, Mishkin, Clark, Krueger, and Sutskever]{radford2021learning}
Alec Radford, Jong~Wook Kim, Chris Hallacy, Aditya Ramesh, Gabriel Goh,
  Sandhini Agarwal, Girish Sastry, Amanda Askell, Pamela Mishkin, Jack Clark,
  Gretchen Krueger, and Ilya Sutskever.
\newblock Learning transferable visual models from natural language
  supervision.
\newblock In \emph{Proceedings of the 38th International Conference on Machine
  Learning}, volume 139 of \emph{Proceedings of Machine Learning Research},
  pages 8748--8763, 2021.

\bibitem[Raghu et~al.(2017)Raghu, Gilmer, Yosinski, and
  Sohl-Dickstein]{raghu2017svcca}
Maithra Raghu, Justin Gilmer, Jason Yosinski, and Jascha Sohl-Dickstein.
\newblock {SVCCA}: Singular vector canonical correlation analysis for deep
  learning dynamics and interpretability.
\newblock In \emph{Advances in Neural Information Processing Systems},
  volume~30, 2017.

\bibitem[Reimers and Gurevych(2019)]{reimers2019sentence}
Nils Reimers and Iryna Gurevych.
\newblock Sentence-{BERT}: Sentence embeddings using {S}iamese {BERT}-networks.
\newblock In \emph{Proceedings of the 2019 Conference on Empirical Methods in
  Natural Language Processing}, pages 3982--3992, 2019.

\bibitem[Sanh et~al.(2019)Sanh, Debut, Chaumond, and Wolf]{sanh2019distilbert}
Victor Sanh, Lysandre Debut, Julien Chaumond, and Thomas Wolf.
\newblock {DistilBERT}, a distilled version of {BERT}: Smaller, faster, cheaper
  and lighter.
\newblock arXiv:1910.01108, 2019.

\bibitem[Schoppe et~al.(2016)Schoppe, Harper, Willmore, King, and
  Schnupp]{schoppe2016measuring}
Oliver Schoppe, Nicol~S. Harper, Ben D.~B. Willmore, Andrew~J. King, and Jan
  W.~H. Schnupp.
\newblock Measuring the performance of neural models.
\newblock \emph{Frontiers in Computational Neuroscience}, 10:\penalty0 10,
  2016.

\bibitem[Song et~al.(2012)Song, Smola, Gretton, Bedo, and
  Borgwardt]{song2012feature}
Le~Song, Alex Smola, Arthur Gretton, Justin Bedo, and Karsten Borgwardt.
\newblock Feature selection via dependence maximization.
\newblock \emph{Journal of Machine Learning Research}, 13:\penalty0 1393--1434,
  2012.

\bibitem[Spearman(1904)]{spearman1904proof}
Charles Spearman.
\newblock The proof and measurement of association between two things.
\newblock \emph{The American Journal of Psychology}, 15\penalty0 (1):\penalty0
  72--101, 1904.

\bibitem[Spearman(1910)]{spearman1910correlation}
Charles Spearman.
\newblock Correlation calculated from faulty data.
\newblock \emph{British Journal of Psychology}, 3\penalty0 (3):\penalty0
  271--295, 1910.

\bibitem[Sucholutsky et~al.(2025)Sucholutsky, Muttenthaler, Weller, Peng, Bobu,
  Kim, Love, Cueva, Grant, Groen, Achterberg, Tenenbaum, Collins, Hermann,
  Oktar, Greff, Hebart, Cloos, Kriegeskorte, Jacoby, Zhang, Marjieh, Geirhos,
  Chen, Kornblith, Rane, Konkle, O'Connell, Unterthiner, Lampinen, M{\"u}ller,
  Toneva, and Griffiths]{sucholutsky2025getting}
Ilia Sucholutsky, Lukas Muttenthaler, Adrian Weller, Andi Peng, Andreea Bobu,
  Been Kim, Bradley~C. Love, Christopher~J. Cueva, Erin Grant, Iris Groen,
  Jascha Achterberg, Joshua~B. Tenenbaum, Katherine~M. Collins, Katherine~L.
  Hermann, Kerem Oktar, Klaus Greff, Martin~N. Hebart, Nathan Cloos, Nikolaus
  Kriegeskorte, Nori Jacoby, Qiuyi Zhang, Raja Marjieh, Robert Geirhos, Sherol
  Chen, Simon Kornblith, Sunayana Rane, Talia Konkle, Thomas~P. O'Connell,
  Thomas Unterthiner, Andrew~K. Lampinen, Klaus-Robert M{\"u}ller, Mariya
  Toneva, and Thomas~L. Griffiths.
\newblock Getting aligned on representational alignment.
\newblock \emph{Transactions on Machine Learning Research}, 2025.

\bibitem[Umar and Laio(2026)]{umar2026snr}
Ali~Hussaini Umar and Alessandro Laio.
\newblock Signal-to-noise ratio and sample size govern representational
  alignment in neural networks.
\newblock arXiv:2605.26973, 2026.

\bibitem[Wang et~al.(2020)Wang, Wei, Dong, Bao, Yang, and Zhou]{wang2020minilm}
Wenhui Wang, Furu Wei, Li~Dong, Hangbo Bao, Nan Yang, and Ming Zhou.
\newblock {MiniLM}: Deep self-attention distillation for task-agnostic
  compression of pre-trained transformers.
\newblock In \emph{Advances in Neural Information Processing Systems},
  volume~33, 2020.

\bibitem[Xu et~al.(2016)Xu, Mei, Yao, and Rui]{xu2016msrvtt}
Jun Xu, Tao Mei, Ting Yao, and Yong Rui.
\newblock {MSR-VTT}: A large video description dataset for bridging video and
  language.
\newblock In \emph{Proceedings of the IEEE Conference on Computer Vision and
  Pattern Recognition}, pages 5288--5296, 2016.

\end{thebibliography}

\end{document}